\documentclass{article}

\usepackage{microtype}
\usepackage{graphicx}
\usepackage{booktabs} %

\usepackage{bbm, dsfont}

\usepackage{amsfonts}
\usepackage{graphicx}
\usepackage{xspace}
\usepackage{xcolor}
\usepackage[colorlinks=true, allcolors=blue]{hyperref}
\usepackage{booktabs}
\usepackage{subcaption}
\usepackage{caption}

\usepackage{hyperref}
\usepackage{url}
\usepackage{listings}
\usepackage{xspace}
\usepackage{array}
\usepackage{makecell}
\usepackage{graphicx}
\usepackage{multirow}
\usepackage{tcolorbox, soul} %
\usepackage{makecell, multicol, color}
\usepackage{booktabs}
\usepackage{subcaption}
\usepackage{wrapfig}
\usepackage{enumitem}
\usepackage{lipsum}
\usepackage{tikz}
\usepackage[stable]{footmisc}

\usepackage[english]{babel}
\addto\extrasenglish{
  
}
\addto\extrasenglish{
  
}

\usepackage{enumitem}
\setitemize{noitemsep,topsep=0pt,parsep=0pt,partopsep=0pt}

\usepackage{amsmath}
\usepackage{amssymb}
\usepackage{mathtools}
\usepackage{amsthm}
\usepackage{subcaption}

\theoremstyle{plain}
\newtheorem{theorem}{Theorem}[section]

\theoremstyle{definition}

\theoremstyle{remark}

\newcommand{\sys}{S-LoRA\xspace}

\newif\ifcomments
\commentsfalse 
\ifcomments
    \newcommand{\todo}[1]{{\color{red}{[TODO: #1]}}}
    \newcommand{\ying}[1]{{\color{blue}{\bf\sf [Ying: #1]}}}
    \newcommand{\shiyi}[1]{{\color{purple}{\bf\sf [Shiyi: #1]}}}
    \newcommand{\ion}[1]{{\color{blue}{\bf\sf [Ion: #1]}}}
    \newcommand{\joey}[1]{{\color{cyan}{\bf\sf [Joey: #1]}}}
    \newcommand{\DL}[1]{{\color{cyan}{\bf\sf [DL: #1]}}}
    \newcommand{\banghua}[1]{{\color{magenta}{\bf\sf [Banghua: #1]}}}
    
    \newcommand{\lisa}[1]{{\color{violet}{\bf\sf [Lisa: #1]}}}
    \newcommand{\nick}[1]{{\color{green}{\bf\sf [Nick: #1]}}}

\else
    \newcommand{\todo}[1]{}
    \newcommand{\ying}[1]{}
    \newcommand{\shiyi}[1]{}
    \newcommand{\ion}[1]{}
    \newcommand{\joey}[1]{}
    \newcommand{\DL}[1]{}
    \newcommand{\banghua}[1]{}
    \newcommand{\lisa}[1]{}
    \newcommand{\nick}[1]{}
\fi

\usepackage{hyperref}

\usepackage[accepted]{mlsys2024}

\mlsystitlerunning{\sys: Serving Thousands of Concurrent LoRA Adapters
}

\begin{document}

\twocolumn[
\mlsystitle{\sys: Serving Thousands of Concurrent LoRA Adapters}

\mlsyssetsymbol{equal}{*}

\begin{mlsysauthorlist}
\mlsysauthor{Ying Sheng}{equal,1,2}
\mlsysauthor{Shiyi Cao}{equal,1}
\mlsysauthor{Dacheng Li}{1}
\mlsysauthor{Coleman Hooper}{1}
\mlsysauthor{Nicholas Lee}{1}
\mlsysauthor{Shuo Yang}{1,3}
\mlsysauthor{Christopher Chou}{1}
\mlsysauthor{Banghua Zhu}{1}
\mlsysauthor{Lianmin Zheng}{1}
\mlsysauthor{Kurt Keutzer}{1}
\mlsysauthor{Joseph E. Gonzalez}{1}
\mlsysauthor{Ion Stoica}{1}
\end{mlsysauthorlist}

\mlsysaffiliation{1}{UC Berkeley}
\mlsysaffiliation{2}{Stanford University}
\mlsysaffiliation{3}{Shanghai Jiao Tong University}

\mlsyscorrespondingauthor{Ying Sheng}{ying1123@stanford.edu}
\mlsyscorrespondingauthor{Shiyi Cao}{shicao@berkeley.edu}

\mlsyskeywords{MLSys，Large Language Models, Low-Rank Adaptation, Multi-tenant Serving}

\vskip 0.3in

\begin{abstract}
The ``pretrain-then-finetune'' paradigm is commonly adopted in the deployment of large language models. Low-Rank Adaptation (LoRA), a parameter-efficient fine-tuning method, is often employed to adapt a base model to a multitude of tasks, resulting in a substantial collection of LoRA adapters derived from one base model.
We observe that this paradigm presents significant opportunities for batched inference during serving. To capitalize on these opportunities, we present \sys, a system designed for the scalable serving of many LoRA adapters.
\sys stores all adapters in the main memory and fetches the adapters used by the currently running queries to the GPU memory.
To efficiently use the GPU memory and reduce fragmentation, \sys proposes Unified Paging.
Unified Paging uses a unified memory pool to manage dynamic adapter weights with different ranks and KV cache tensors with varying sequence lengths.
Additionally, \sys employs a novel tensor parallelism strategy and highly optimized custom CUDA kernels for heterogeneous batching of LoRA computation.
Collectively, these features enable \sys to serve thousands of LoRA adapters on a single GPU or across multiple GPUs with a small overhead.
Compared to state-of-the-art libraries such as HuggingFace PEFT and vLLM (with naive support of LoRA serving), \sys can improve the throughput by up to 4 times and increase the number of served adapters by several orders of magnitude. As a result, \sys enables scalable serving of many task-specific fine-tuned models and offers the potential for large-scale customized fine-tuning services.
The code is available at \url{https://github.com/S-LoRA/S-LoRA}.
\end{abstract}
]

\printAffiliationsAndNotice{\mlsysEqualContribution} %

\section{Introduction}
\label{sec:intro}

Large language models (LLMs) have become ubiquitous in modern applications, ranging from natural language processing to more general tasks~\cite{openai2023gpt4, touvron2023llama2, alayrac2022flamingo}.
Within these domains, LLMs have consistently demonstrated superior performance, especially when fine-tuned for specific tasks~\cite{kenton2019bert, houlsby2019parameter, ouyang2022training}.
This ``pretrain-then-finetune'' paradigm has led to the proliferation of numerous fine-tuned variants of a single base LLM, each tailored to a specific task or domain.

When scaling the fine-tuning of a base model for numerous tasks, such as personalized assistants, which could involve thousands or millions of users, the associated training and serving costs can become substantial.
To address this, several parameter-efficient fine-tuning methods have been developed. 
A prime exemplar is Low-Rank Adaptation (LoRA)~\cite{hu2021lora}, which enables efficient fine-tuning by updating only low-rank additive matrices.
These matrices consist of a small number of parameters, referred to as adapter weights.
LoRA has shown that by fine-tuning just these adapter weights, it is possible to achieve performance on par with full-weight fine-tuning. However, despite considerable research into fine-tuning, the question of how to serve these fine-tuned variants at scale remains unexplored.

One of the key innovations in the LoRA paper was the elimination of adapter inference latency by directly merging the adapter with the model parameters. Additionally, to support multiple models on a single machine, the same paper proposes swapping adapters by adding and subtracting LoRA weights from the base model. While this approach enables low-latency inference for a single adapter and serial execution across adapters, it significantly reduces overall serving throughput and increases total latency when serving multiple adapters concurrently. Moreover, the paper does not consider the opportunity to leverage host memory to increase the number of adapters hosted by a single machine.
\DL{Moreover, the paper simply places adapters in the device memory, and forgoes more sophisticated memory allocation strategy, e.g. using the much larger host memory to increase the number of adapters hosted by a single machine.}

In this paper, we study how to scalably serve thousands of LoRA adapters on a single machine.
We observe that the shared base model, which underpins numerous LoRA adapters, presents a substantial opportunity for batched inference. To achieve high-throughput multi-adapter serving, it is advantageous to separate the batchable base model computation from individual LoRA computations.

While leveraging batching in the base model is straightforward (as all queries share the base model), extending batching to the adapters is challenging.
First, serving many LoRA adapters simultaneously requires efficient memory management. Since GPU memory is limited, we must store adapter weights outside the GPU and dynamically fetch them when needed. However, dynamically loading and unloading adapters of varying sizes, coupled with the dynamic allocation and deallocation of KV cache tensors for requests with different sequence lengths, can lead to significant memory fragmentation and I/O overhead.
Second, apart from the easily batchable base model computation, the separated computation of many adapters with distinct ranks in non-contiguous memory is challenging to batch and demands the development of new computation kernels.
Third, leveraging multiple GPUs on a single machine requires novel parallelism strategies to accommodate the added LoRA weights and computations. It is essential to carefully design this strategy to minimize communication and memory overheads.

To this end, we introduce \sys, a scalable LoRA serving system.
\sys exploits batching opportunities, efficiently manages both host and GPU memory, and orchestrates parallelism across multiple GPUs.
The primary contributions of \sys are summarized as follows: 

\begin{itemize}
    \item \textit{Unified Paging}: To reduce memory fragmentation and increase batch size, \sys introduces a unified memory pool. This pool manages dynamic adapter weights and KV cache tensors by a unified paging mechanism.
    \item \textit{Heterogeneous Batching}: To minimize the latency overhead when batching different adapters of varying ranks, \sys employs highly optimized custom CUDA kernels. These kernels operate directly on non-contiguous memory and align with the memory pool design, facilitating efficient batched inference for LoRA.
    \item \textit{\sys TP}: To ensure effective parallelization across multiple GPUs, \sys introduces a novel tensor parallelism strategy. This approach incurs minimal communication cost for the added LoRA computation compared to that of the base model. This is realized by scheduling communications on small intermediate tensors and fusing the large ones with the communications of the base model.
\end{itemize}

We evaluate \sys by serving Llama-7B/13B/30B/70B.
Results show that \sys can serve thousands of LoRA adapters on a single GPU or across multiple GPUs with a small overhead.
When compared to the state-of-the-art parameter-efficient fine-tuning library, Huggingface PEFT, \sys can enhance throughput by up to $30\times$. In comparison to the high-throughput serving system vLLM using a naive support of LoRA serving, \sys can improve throughput by up to $4 \times$ and increase the number of served adapters by several orders of magnitude.

\section{Background}
\label{sec:background}

Low-Rank Adaptation (LoRA)~\cite{hu2021lora} is a parameter-efficient fine-tuning method designed to adapt pre-trained large language models to new tasks. The motivation behind LoRA stems from the low intrinsic dimensionality of model updates during adaptation.
In the training phase, LoRA freezes the weights of a pre-trained base model and adds trainable low-rank matrices to each layer. This approach significantly reduces the number of trainable parameters and memory consumption. When compared to full parameter fine-tuning, LoRA can often reduce the number of trainable parameters by orders of magnitude (e.g., $10000\times$) while retaining comparable accuracy.
For the inference phase, the original paper suggests merging the low-rank matrices with the weights of the base model. As a result, there is no added overhead during inference, setting it apart from previous adapters like~\cite{houlsby2019parameter} or prompt tuning methods such as~\cite{lester2021power}.

Formally, for a pre-trained weight matrix $W \in \mathbb{R}^{h \times d}$, LoRA introduces the update as $W^\prime = W + AB$, where $A \in \mathbb{R}^{h \times r}$, $B \in \mathbb{R}^{r \times d}$, and the rank $r \ll \min(h, d)$.
If the forward pass of a base model is defined by $h = xW$, then after applying LoRA, the forward pass becomes 
\begin{align}
    h& =  x W^\prime = x (W + AB) \label{eq:lora} \\
    & = xW + xAB  \label{eq:lora_factored}.
\end{align}
Typically, this adjustment is only applied to the query, key, value, and output projection matrices in the self-attention module, excluding the feed-forward module.

Because LoRA greatly reduces the training and weight storage costs, it has been widely adopted by the community, and people have created hundreds of thousands of LoRA adapters for pre-trained large language models and diffusion models~\cite{peft}.

\subsection{Serving Large Language Models}
Most large language models (LLMs) are based on the transformer architecture~\cite{vaswani2017attention}.
The number of parameters in an LLM ranges from several billion to several trillion~\cite{brown2020language,chowdhery2022palm,fedus2022switch}, corresponding to disk sizes spanning several gigabytes to even terabytes.
This scale results in LLM serving having significant computational and memory demands.

Additionally, the inference process for LLMs requires iterative autoregressive decoding.
Initially, the model carries out a forward pass to encode the prompt. Following this, it decodes the output one token at a time. The sequential process makes decoding slow.
Since each token attends to the hidden states of all its preceding tokens, it becomes essential to store the hidden states of all previous tokens. This storage is referred to as the ``KV cache''.
Such a mechanism adds to the memory overhead and causes the decoding process to be more memory-intensive than computation-intensive.

The challenges become even more pronounced in online settings, where requests of varying sequence lengths arrive dynamically.
To accommodate such dynamic incoming requests, Orca~\cite{yu2022orca} introduces a method of fine-grained, iteration-level scheduling. Instead of scheduling at the request level, Orca batches at the token level. This approach allows for the continuous addition of new requests to the currently running batch, resulting in substantially higher throughput.
vLLM~\cite{kwon2023efficient} further optimizes Orca's memory efficiency using PagedAttention.
PagedAttention adopts concepts from virtual memory and paging in operating systems and manages the storage and access of dynamic KV cache tensors in a paged fashion. This method efficiently reduces fragmentation, facilitating larger batch sizes and higher throughput.

When serving very large models that exceed the memory capacity of a single GPU, or when there are stringent latency requirements, it is necessary to parallelize the model across multiple GPUs. Several model parallelism methods have been proposed, such as tensor parallelism~\cite{shoeybi2019megatron}, sequence parallelism~\cite{korthikanti2023reducing}, pipeline parallelism~\cite{huang2019gpipe}, and their combinations~\cite{narayanan2021efficient, zheng2022alpa}.
\ying{explain metrics}

\section{Overview of \sys}
\sys encompasses three principal components of innovation.
In \autoref{sec:scheduling}, we introduce our batching strategy, which decomposes the computation between the base model and the LoRA adapters.
Additionally, we discuss adapter clustering and admission control when scheduling the requests.
The ability to batch across concurrent adapters, introduces new challenges around memory management.
In \autoref{sec:memory-management}, we generalize PagedAttention~\cite{kwon2023efficient} to Unfied Paging, which supports dynamically loading LoRA adapters. 
This approach uses a unified memory pool to store the KV caches and adapter weights in a paged fashion, which can reduce fragmentation and balance the dynamic changing size of the KV caches and adapter weights.
In \autoref{sec:tensor-parallelism}, we introduce our new tensor parallelism strategy that enables us to efficiently decouple the base model and LoRA adapters.

\section{Batching and Scheduling}
\label{sec:scheduling}

\subsection{Batching}

\begin{figure}[t]
    \centering
    \includegraphics[width=0.85\linewidth]{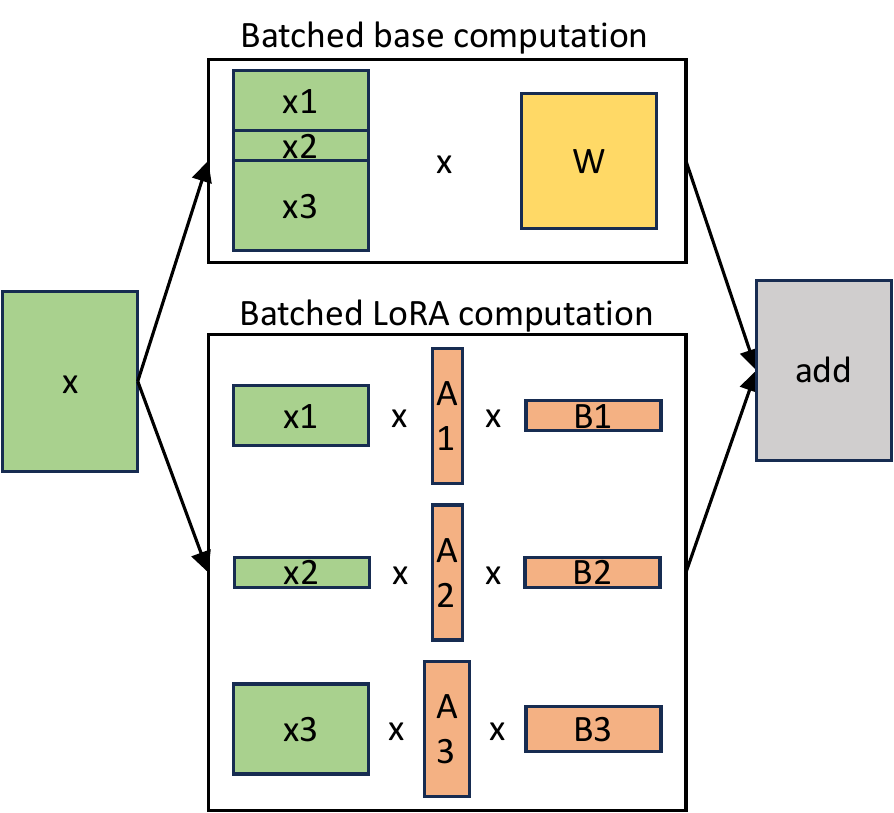}
    \caption{Separated batched computation for the base model and LoRA computation. The batched computation of the base model is implemented by GEMM. The batched computation for LoRA adapters is implemented by custom CUDA kernels which support batching various sequence lengths and adapter ranks.}
    \label{fig:batch_base}
\end{figure}

\begin{figure}[t]
    \centering
    \includegraphics[width=\linewidth]{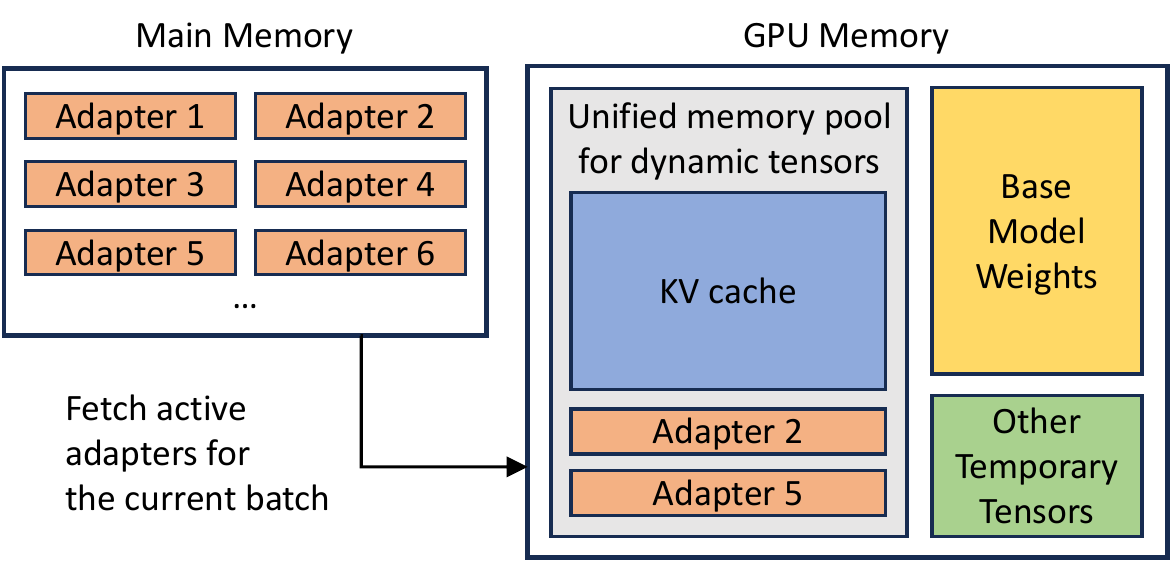}
    \caption{Overview of memory allocation in \sys. \sys stores all adapters in the main memory and fetches the active adapters for the current batch to the GPU memory. The GPU memory is used to store the KV cache, adapter weights, base model weights, and other temporary tensors.}
    \label{fig:storage_overview}
\end{figure}

Our batching strategy aims to support online and high-throughput serving of many LoRA adapters simultaneously.

For a single adapter, the method recommended by \cite{hu2021lora} is to merge the adapter weights into the base model weights, resulting in a new model (see Eq.~\ref{eq:lora}).
This has the advantage that there is no additional adapter overhead during inference, since the new model has the same number of parameters as the base model. 
In fact, this was a prominent feature of the original LoRA work.

However, when there are multiple adapters, merging the weights into the base model leads to multiple weight copies and missed batching opportunities.
Directly merging the models requires maintaining many copies of the full language model. 
In the original LoRA paper, the authors proposed adding and subtracting LoRA weights on the fly to enable serving multiple models without increasing the memory overhead.  
However, this approach doesn't support concurrent inference on separate LoRA adapters and therefore limits batching opportunities.  

In this paper, we show that merging LoRA adapters into the base model is inefficient for the multi-LoRA high-throughput serving setting.
Instead, we propose computing the  LoRA computation $xAB$ on-the-fly as shown in Eq.~\ref{eq:lora_factored}.
This avoids weight duplication and enables batching of the more costly $xW$ operation.
But this approach also increases the computation overhead.
However, because the cost of $xAB$ is substantially lower than $xW$ and there is a considerable savings from batching $xW$ across different adapters, we show that the savings far exceed the additional overhead.

Unfortunately, directly implementing the factored computation of the base model and individual LoRA adapters using the batch GEMM kernel from the existing BLAS libraries would require significant padding and result in poor hardware utilization.
This is because of the \textit{heterogeneity} of sequence lengths and adapter ranks. 

In \sys, we batch the computation of the base model and then employ custom CUDA kernels to execute the additional $xAB$ for all adapters separately.
This process is illustrated by \autoref{fig:batch_base}.
Instead of naively using padding and using the batch GEMM kernel from the BLAS library for the LoRA computation, we implement custom CUDA kernels for more efficient computation without padding.
In \autoref{subsec:custom-kernel}, we discuss the implementation details.

While the number of LoRA adapters can be large if we store them in main memory, the number of LoRA adapters needed for the currently running batch is manageable, because the batch size is bounded by the GPU memory.
To take advantage of this, we store all LoRA adapters in the main memory and fetch only the LoRA adapters needed for the currently running batch to the GPU RAM when running the inference for that batch.
In this case, the maximum number of adapters that can be served is bounded by the main memory size.
This process is illustrated by \autoref{fig:storage_overview}.
To achieve high-throughput serving, we adopt the iteration-level scheduling batching strategy from Orca~\cite{yu2022orca}. In this approach, requests are scheduled at the token level.
We immediately incorporate a new request into the running batch if space is available. The request will exit the batch once it reaches the maximum number of generated tokens or fulfills other stopping criteria.
This process reduces GPU memory usage but introduces new memory management challenges. In \autoref{sec:memory-management}, we will discuss our techniques to manage memory efficiently.

\subsection{Adapter Clustering}
To enhance batching efficiency, one potential strategy is reducing the number of active adapters in a running batch. By using fewer adapters, there is an opportunity to allocate more memory to the KV cache, which in turn can facilitate larger batch sizes. 
Given the common memory capacities of GPUs, they are often underutilized while decoding. Consequently, increasing the batch size can lead to higher throughput.
A direct approach to reducing the number of adapters in a running batch is to prioritize batching requests that use the same adapter, a strategy we term ``adapter clustering". However, adapter clustering comes with its own set of trade-offs. For example, it can hurt the average latency or fairness among adapters. We provide an ablation study in \autoref{sec:full_exp} to illustrate how throughput and latency change according to the cluster size.

\subsection{Admission Control}
\label{subsec:admission-control}
\joey{this sort of comes out of nowhere and needs more motivation for the problem.  Why is admission control needed for LoRA serving and what is special about it?  Shouldn't the clustering of adapters cause me to preferentially admit queries that use adapters that are already in the queue?}
\nick{I think that one could move the introduction to an admission control can be moved to the Background, as it is literally just background information. You only need to discuss the reward function and the theorem in relation to the SLO attainment metric in section 4.1.}
In \sys, we also applied an admission control strategy to sustain good attainment when the traffic is higher than the serving system capacity.
A serving system is typically characterized by a service level objective (SLO) which specifies the desired latency of processing requests. If the serving system has fixed capacity, it must implement an admission control mechanism, that drops a request, if the system cannot meet its SLO. Otherwise, if no request is dropped, and the number of incoming requests is larger than the system capacity for long enough, the serving system is bound to violate the SLO.
We implemented an abort strategy to mimic admission control in \sys, called early abort strategy.
Intuitively, we estimate the set of latest requests that we can serve in SLO, and then serve them in the order of arrival time.
More implementation details and mathematical justifications are deferred to \autoref{sec:admission-detail}.

\section{Memory Management}
\label{sec:memory-management}

Compared to serving a single base model, serving multiple LoRA adapters simultaneously presents new memory management challenges.
To support many adapters, \sys stores them in the main memory and dynamically loads the adapter weights needed for the currently running batch into GPU RAM.
During this process, there are two noticeable challenges. The first is memory fragmentation, resulting from the dynamic loading and offloading adapter weights of various sizes. The second is the latency overhead introduced by adapter loading and offloading.
To tackle these challenges efficiently, we propose Unfied Paging and overlap the I/O with computation by prefetching adapter weights.

\subsection{Unified Paging}
Understanding the nature of adapter weights is essential for optimizing memory usage. Our primary observation is that these dynamic adapter weights are analogous to dynamic KV caches in several ways:

\begin{itemize} 
\item \textbf{Variable sizes and operations:} 
Just as the size of KV cache size fluctuates with the sequence length, the ranks of the active adapters can also depend on the choice of adapter associated with each request. 
KV caches are allocated when requests arrive and deallocated once the requests are completed. 
Similarly, adapter weights 
are loaded and cleared with each request. If not managed properly, this variability can result in fragmentation.

\item \textbf{Dimensionality:} A KV cache tensor for a request in a layer has a shape of $(S, H)$, where $S$ denotes the sequence length and $H$ represents the hidden dimension. Meanwhile, the shape of a LoRA weight is $(R, H)$, with $R$ standing for the rank and $H$ the hidden dimension. Both share a dimension size of $H$ 
that can be leveraged to reduce fragmentation. 
\end{itemize}

\begin{figure}
    \centering
    \includegraphics[width=0.7\linewidth]{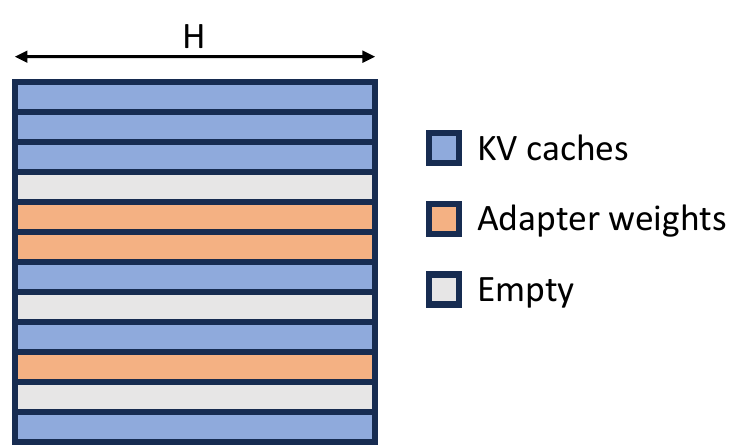}
    \vspace{-1em}
    \caption{Unified memory pool. We use a unified memory pool to store both KV caches and adapter weights in a non-contiguous way to reduce memory fragmentation. The page size is $H$ elements.}
    \label{fig:unified_memory_pool}
\end{figure}

Motivated  by these parallels, we extend the idea of PagedAttention~\cite{kwon2023efficient} to Unified Paging which manages adapter weights in addition to the KV cache.
Unified Paging uses a unified memory pool to jointly manage both KV cache and adapter weights.
To implement this, we first allocate a large buffer statically for the memory pool. 
This buffer uses all available space except for the space occupied by the base model weights and temporary activation tensors.
Both KV caches and adapter weights are stored in this memory pool in a paged manner, with each page corresponding to a vector of $H$.
Thus, a KV cache tensor with a sequence length of $S$ uses up $S$ pages, while a LoRA weight tensor of rank $R$ takes up $R$ pages.
\autoref{fig:unified_memory_pool} illustrates the layout of our memory pool, where KV caches and adapter weights are stored interleaved and non-contiguously. This approach significantly reduces fragmentation, ensuring that adapters weights of various ranks can coexist with dynamic KV caches in a structured and systematic manner.

\subsection{Prefetching and Overlapping}
Although the unified memory pool mitigates fragmentation, the I/O overhead from loading and offloading remains a concern—especially when dealing with numerous or large adapters. The latency introduced by waiting to load these adapters can compromise the efficiency of the system.

To proactively address this issue, we introduce a dynamic prediction mechanism. While running the current decoding batch, we predict the adapters required for the next batch based on the current waiting queue. This prediction allows us to prefetch and store them in available memory. Such a forward-looking strategy keeps most of the adapters needed for the next batch already in place before running it, which reduces I/O time for adapter swapping.

\subsection{Custom Kernels for heterogeneous LoRA batching on Non-Contiguous Memory}
\label{subsec:custom-kernel}
Due to the design of the unified memory pool, the adapter weights are stored in non-contiguous memory.
To run computations efficiently under this design, we implement custom CUDA kernels that support batching LoRA computations with \emph{varying ranks} and sequence lengths in a \emph{non-contiguous} memory layout.

In the prefill stage, the kernel handles a sequence of tokens and gathers adapter weights with different ranks from the memory pool.
We call this kernel Multi-size Batched Gather Matrix-Matrix Multiplication (MBGMM). It is implemented in Triton~\cite{tillet2019triton} with tiling.

In the decode stage, the kernel handles a single token and gathers adapter weights with different ranks from the memory pool.
We call this kernel Multi-size Batched Gather Matrix-Vector Multiplication (MBGMV).
We implemented two versions of this kernel: one in Triton and another by modifying an earlier version of Punica kernels~\cite{punica} to extend support for non-contiguous memory, multiple ranks in a batch, and more fine-grained memory gathering. We found the latter one was faster, so we used it in the experiments.

Punica~\cite{chen2023punica} is concurrent work on serving multiple LoRA adapters, which will be discussed in Section~\ref{sec:related-work}. In addition to Triton and Pucina kernels, NVIDIA CUTLASS also provides high-performance kernels for grouped GEMM~\cite{cutlass_grouped_gemm} that can be used for heterogeneous batching.

\section{Tensor Parallelism}
\label{sec:tensor-parallelism}

We design novel tensor parallelism strategies for batched LoRA inference to support multi-GPU inference of large transformer models.
Tensor parallelism is the most widely used parallelism method because its single-program multiple-data pattern simplifies its implementation and integration with existing systems.
Tensor parallelism can reduce the per-GPU memory usage and latency when serving large models.
In our setting, the additional LoRA adapters introduce new weight matrices and matrix multiplications, which calls for new partition strategies for these added items.

\begin{figure*}[ht]
    \centering
    \includegraphics[width=0.95\linewidth]{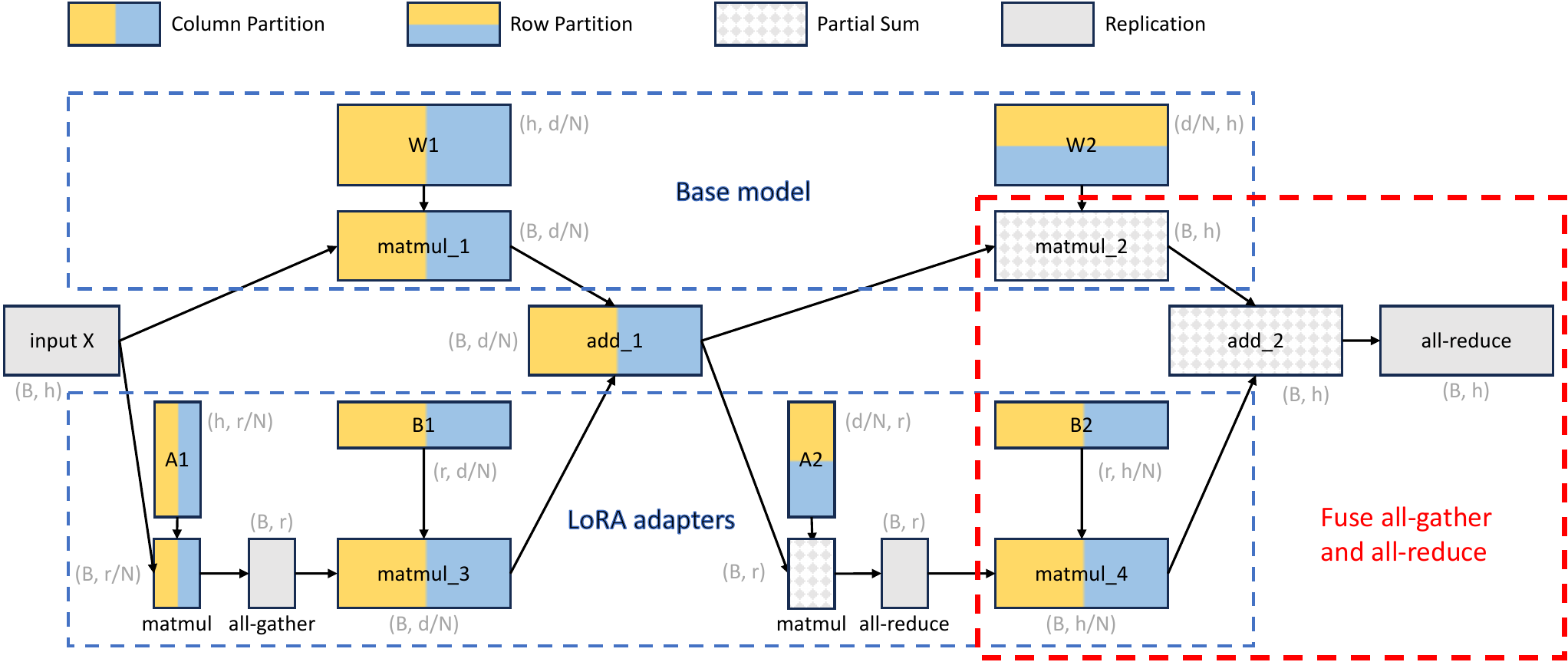}
    \vspace{-1em}
    \caption{Tensor parallelism partition strategy for batched LoRA computation. This is a computational graph where nodes represent tensors/operators and the edges represent dependency. We use different colors to represent different partition strategies, which include column partition, row partition, partial sum, and replication. The per-GPU shape of each tensor is also annotated in gray.
    Note that $B$ is the number of tokens, $h$ is the input dimension, $N$ is the number of devices, $d$ is the hidden size, and $r$ is the adapter rank.
    }
    \label{fig:lora_tp}
\end{figure*}

\subsection{Partition Strategy}
\label{subsec:partition-strategy}
Since the base model uses the Megatron-LM tensor parallelism strategy~\cite{shoeybi2019megatron}, our approach aims to align the partition strategies of inputs and outputs of the added LoRA computation with those of the base model.
In this way, we can minimize the communication costs by avoiding unnecessary communications and fusing some communications.

We use the feed-forward module (2-layer MLP) to illustrate our partition strategy. We will explain later how this strategy can easily be adapted to the self-attention layer. As depicted in \autoref{fig:lora_tp}, the upper box illustrates the base model's Megatron-LM partition strategy: the first weight matrix ($W1$) is column-partitioned, and the second ($W2$) is row-partitioned. An all-reduce communication is required to accumulate the partial sum from distributed devices.

The lower box illustrates the partitioning strategy for the added LoRA computation. The matrices $A1$ and $B1$ for the adapter of the first weight matrix ($W1$) are column-partitioned. An all-gather operation is used to collect the intermediate results.
The matrices $A2$ and $B2$ for the adapter of the second weight ($W2$) are row-partitioned and column-partitioned, respectively.
An all-reduce operation is used to sum up the intermediate results. Finally, the result from the LoRA computation is added to that from the base model (\texttt{add\_2}).
A single all-reduce operation is sufficient to accumulate the final results. It is worth noting that we are essentially fusing an all-gather operation for \texttt{matmul\_4} with the final all-reduce.
To our knowledge, this parallelization strategy has not been studied before.

Next, we discuss adapting the strategy from the 2-layer MLP to the self-attention layer.
Similar to the Megatron-LM strategy, we partition the head dimension of the self-attention layer.
The query-key-value projection weight matrix can be seen as $W1$ in our example and the output projection weight matrix can be seen as $W2$ in our example.

\subsection{Communication and Memory Cost Analysis}
\label{subsec:communication-cost-analysis}

Let $N$ be the number of devices, $B$ be the number of tokens, $h$ be the hidden size, and $r$ be the adapter rank.
The communication cost of the base model is one all-reduce, or $\frac{2(N-1)Bh}{N}$. The communication cost of the added LoRA computation is three all-gather for query, key, and value projections, and one all-reduce for the output projection. Formally, it is $3 \frac{(N-1)Br}{N} + \frac{2(N-1)Br}{N} = \frac{5(N-1)Br}{N}$.

Under our strategy, the additional communication cost introduced by LoRA is negligible when compared to the communication cost of the base model, because $r \ll h$.
Intuitively, this is achieved by carefully scheduling communications on the small intermediate tensors of LoRA computation and fusing communications with base models.

In terms of memory usage, our strategy is optimal because we partition all weight matrices among all devices and there is no replicated weight matrix.

\section{Evaluation}
\label{sec:evaluation}

We evaluate the performance of \sys on both synthetic and real production workloads. \sys is built on top of LightLLM~\cite{lightllm}, a single-model LLM serving system based on PyTorch~\cite{paszke2019pytorch} and Triton~\cite{tillet2019triton}.
We evaluate the scalability of \sys by serving up to two thousand LoRA adapters simultaneously and compare it with other strong baselines.
We then perform ablation studies to verify the effectiveness of individual components.

\subsection{Setup}

\textbf{Model.}
We test the Llama model series~\cite{touvron2023llama,touvron2023llama2}, one of the most popular open large language models.
We consider 5 different model and adapter configurations, which are listed in \autoref{tab:model_setting}
\footnote{For Llama-70B, we used different architecture parameters than the official model and did not employ group-query attention.}.
Our optimizations can be easily adapted to other transformer-based architectures as well, such as GPT-3~\cite{brown2020language} and PaLM~\cite{chowdhery2022palm,anil2023palm}.

\begin{table}[ht]
\centering
\footnotesize
\begin{tabular}{cccl}
\toprule
Setting & Base model & Hidden size & Adapter ranks\\
\midrule
S1 & Llama-7B  & 4096 & \{8\} \\
S2 & Llama-7B  & 4096 & \{64, 32, 16, 8\} \\
\midrule
S4 & Llama-13B & 5120 & \{64, 32, 16\} \\
\midrule
S5 & Llama-30B & 7168 & \{32\} \\
S6 & Llama-70B & 8192 & \{64\} \\
\bottomrule
\end{tabular}
\caption{Model and adapter configurations.}
\label{tab:model_setting}
\vspace{-1em}
\end{table}

\textbf{Hardware.}
    We conduct tests on various hardware settings, including a single NVIDIA A10G GPU (24GB), a single A100 GPU (40GB), a single A100 GPU (80GB), and multiple A100 GPUs (40GB/80GB).
The host's main memory varies based on the GPU setup, ranging from 64 GB to 670 GB. We will show that \sys can efficiently scale the number of adapters, limited only by the available main memory.

\textbf{Baselines.}
We benchmark several variants of \sys, HuggingFace PEFT~\cite{peft}, and vLLM~\cite{kwon2023efficient}.

\begin{itemize}
    \item ``HuggingFace PEFT'' is a library for training and running parameter-efficient fine-tuning models. It lacks advanced batching and memory management. We build a server using it that batches single adapter requests and switches adapter weights between batches.
    \item ``vLLM $m$-packed'' is a simple multi-model serving solution based on vLLM, a high-throughput serving system. Because vLLM does not support LoRA, we merge the LoRA weights into the base model and serve the multiple versions of the merged weights separately. To serve $m$ LoRA adapters, we run $m$ vLLM workers on a single GPU, where multiple workers are separate processes managed by NVIDIA MPS.
    We statistically allocate the GPU memory proportionally to the average request rate for each process.
    \item ``\sys'' is \sys with all the optimizations and it is using the first-come-first-serve scheduling strategy.
    \item ``\sys-no-unify-mem'' is \sys without the unified memory management.
    \item ``\sys-bmm'' is \sys without unified memory management and customized kernels. It copies the adapter weights to continuous memory space and performs batched matrix multiplication with padding.

\end{itemize}

\textbf{Metrics.}
There are several metrics to measure the performance of serving systems, including latency and throughput.
Following common practice, we report the \textit{throughput, average request latency, average first token latency, and SLO attainment}.
SLO attainment is defined as the percentage of requests that return the first token in 6 seconds. 
Additionally, we introduce a new metric termed \textit{user satisfaction} (see \autoref{sec:admission-detail}), which offers a more fine-grained analysis of the first token latency.
Intuitively, a shorter first token latency gives a higher satisfaction. The satisfaction becomes 0 if the first token latency exceeds the SLO.

\subsection{End-to-End Results on Synthetic Workloads}

\textbf{Workload trace.} We generate synthetic workload traces using the Gamma process, which is commonly used in machine learning serving literature~\cite{crankshaw2020inferline, li2023alpaserve}.
Given $n$ adapters, the requests for adapter $i$ are modeled using a Gamma arrival process with a mean rate of $\lambda_i$ and a coefficient of variance (CV) of $cv$.
The mean rate, $\lambda_i$, adheres to a power-law distribution with an exponent $\alpha$. The total request rate for all adapters is $R$ requests per second.
For the $n$ adapters, we set their ranks based on the list provided in \autoref{tab:model_setting} with a round-robin method.
Our tests cover various combinations of $n$, $\alpha$, $R$, and $cv$.
For every request, the input and output lengths are sampled from uniform distributions $U[I_l, I_u]$ and $U[O_l, O_u]$ respectively.
The default duration of a trace is 5 minutes.
To conduct comprehensive experiments, we first pick a set of default parameters for generating workloads, as shown in \autoref{tab:default_trace}. We then vary one of the \textit{$n$, $\alpha$, $R$, and $cv$} to see how each factor affects the performance.

\begin{table}[ht]
\caption{Default parameters for generating the synthetic workloads. ``7B @ A10G'' means running a Llama-7B on a single A10G.}
\centering
\resizebox{1\columnwidth}{!}{
\begin{tabular}{c|cccccc}
\toprule
Setting & $n$ & $\alpha$ & $R$ & $cv$ & [$I_l, I_u$] & [$O_l, O_u$] \\
\midrule
7B @ A10G (24G)  & 200 & 1 & 2 & 1 & [8, 512] & [8, 512]\\
7B @ A100 (80G)  & 200 & 1 & 10 & 1 & [8, 512] & [8, 512]\\
13B @ A100 (40G) & 200 & 1 & 2  & 1 & [8, 512] & [8, 512]\\
13B @ A100 (80G) & 400 & 1 & 6 & 1 & [8, 512] & [8, 512]\\
\bottomrule
\end{tabular}}
\label{tab:default_trace}
\vspace{-1em}
\end{table}

\begin{table}[ht]
\centering
\caption{Throughput (req/s) comparison between \sys, vLLM-packed, and PEFT. The hardware is a single A100 (80GB).
We run PEFT for a shorter duration when $n=100$.
We do not evaluate PEFT for $n\geq 1000$, as its throughput is already very low for a small $n$. ``OOM'' denotes out-of-memory.}
\resizebox{0.98\columnwidth}{!}{
\begin{tabular}{c|c|ccc}
\toprule
Model Setup & $n$ & \sys & vLLM-packed & PEFT\\
\midrule
\multirow{4}{*}{S1} & 5 & 8.05 & 2.04 & 0.88 \\
 & 100 & 7.99 & OOM & 0.25\\
 & 1000 & 7.64 & OOM & - \\
 & 2000 & 7.61 & OOM & - \\
\midrule
\multirow{4}{*}{S2} & 5 & 7.48 & 2.04 & 0.74\\
 & 100 & 7.29 & OOM & 0.24 \\
 & 1000 & 6.69 & OOM & -\\
 & 2000 & 6.71 & OOM & -\\
\midrule
\multirow{3}{*}{S4} & 2 & 4.49 & 3.83 & 0.54\\
 & 100 & 4.28 & OOM & 0.13 \\
 & 1000 & 3.96 & OOM & - \\
\bottomrule
\end{tabular}
}
\label{table:vs_vllm_peft}
\end{table}

\begin{figure*}[ht]
    \centering
    \includegraphics[width=\textwidth]{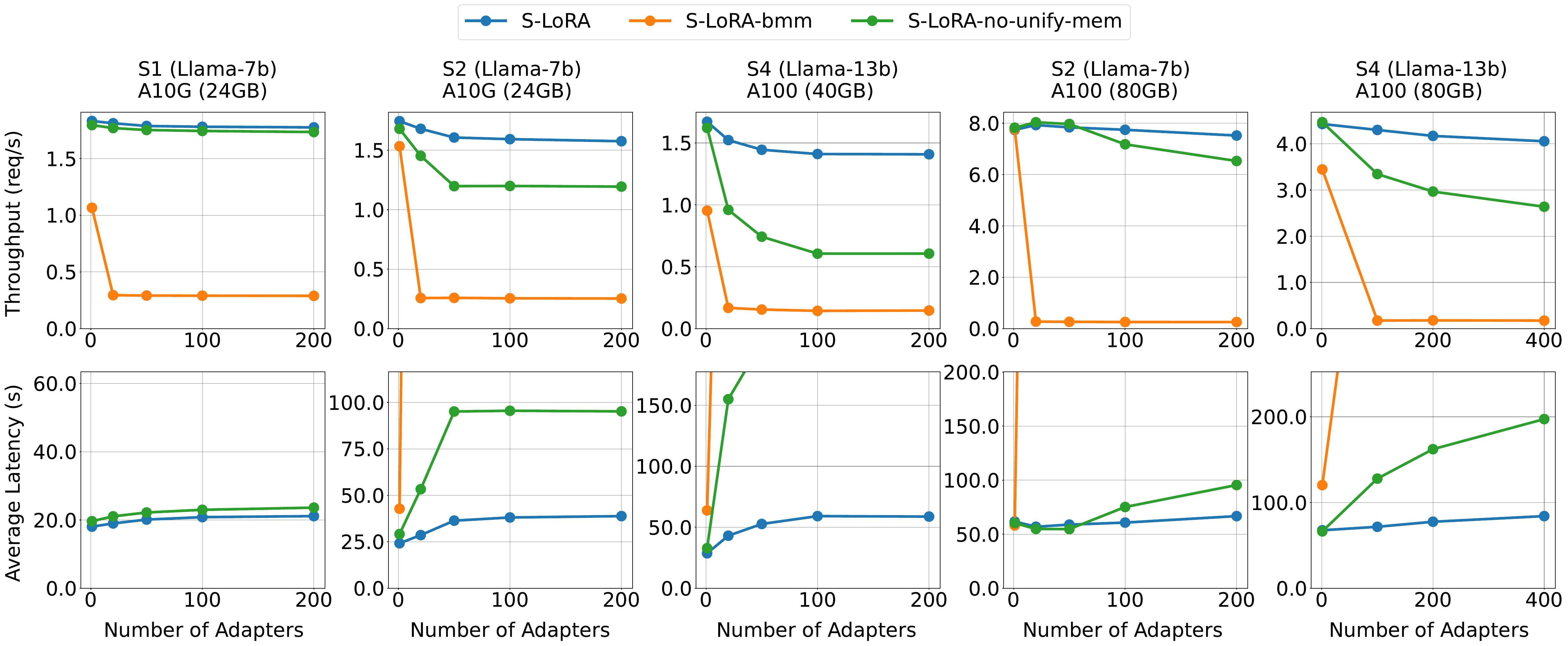}
    \caption{The throughput and average request latency of \sys and its variants under different numbers of adapters. \sys achieves significantly better performance and can scale to a large number of adapters. We run \sys-bmm for a shorter duration since it has a significantly lower throughput.
    Some \sys-bmm curves are omitted because it is out of the figures's scope.}
    \label{fig:main_num_adapter}
    \vspace{-1em}
\end{figure*}

\textbf{Comparison with other systems.}
We compare \sys with both vLLM-packed and HuggingFace PEFT for serving many LoRA adapters. The results are shown in \autoref{table:vs_vllm_peft}.
Remarkably, \sys can serve 2,000 adapters simultaneously, maintaining minimal overhead for the added LoRA computation.
In contrast, vLLM-packed needs to maintain multiple weight copies and can only serve fewer than 5 adapters due to the GPU memory constraint.
The throughput of vLLM-packed is also much lower due to the missed batching opportunity.
Although PEFT can swap adapters between batches, enabling it to handle a large number of adapters, its lack of advanced batching methods and memory management results in significantly worse performance.
Overall, \sys achieves a throughput up to 4x higher than vLLM-packed when serving a small number of adapters, and up to 30x higher than PEFT, while supporting a significantly larger number of adapters.

\textbf{Comparing with own variants.}
Since no baseline system can efficiently scale to a large number of adapters, we now focus on comparing \sys with its own variants.
\autoref{fig:main_num_adapter} illustrates how they scale with the number of adapters. \sys achieves noticeably higher throughput and lower latency compared to \sys-bmm and \sys-no-unify-mem. This implies that our memory pool and custom kernels are effective. When the number of adapters increases, the throughput of \sys initially experiences a slight decline due to the overhead introduced by LoRA. However, once the number of adapters reaches a certain threshold (e.g., 100 in most experiments), the throughput of \sys no longer decreases.
This stability can be attributed to the fact that as the number of adapters grows, the number of activated adapters for the currently running batch remains unchanged, maintaining a constant overhead.
Consequently, \sys can scale to a much larger number of adapters without incurring additional overhead, constrained only by the available main memory.

\begin{figure}[ht]
    \includegraphics[width=0.48\textwidth]{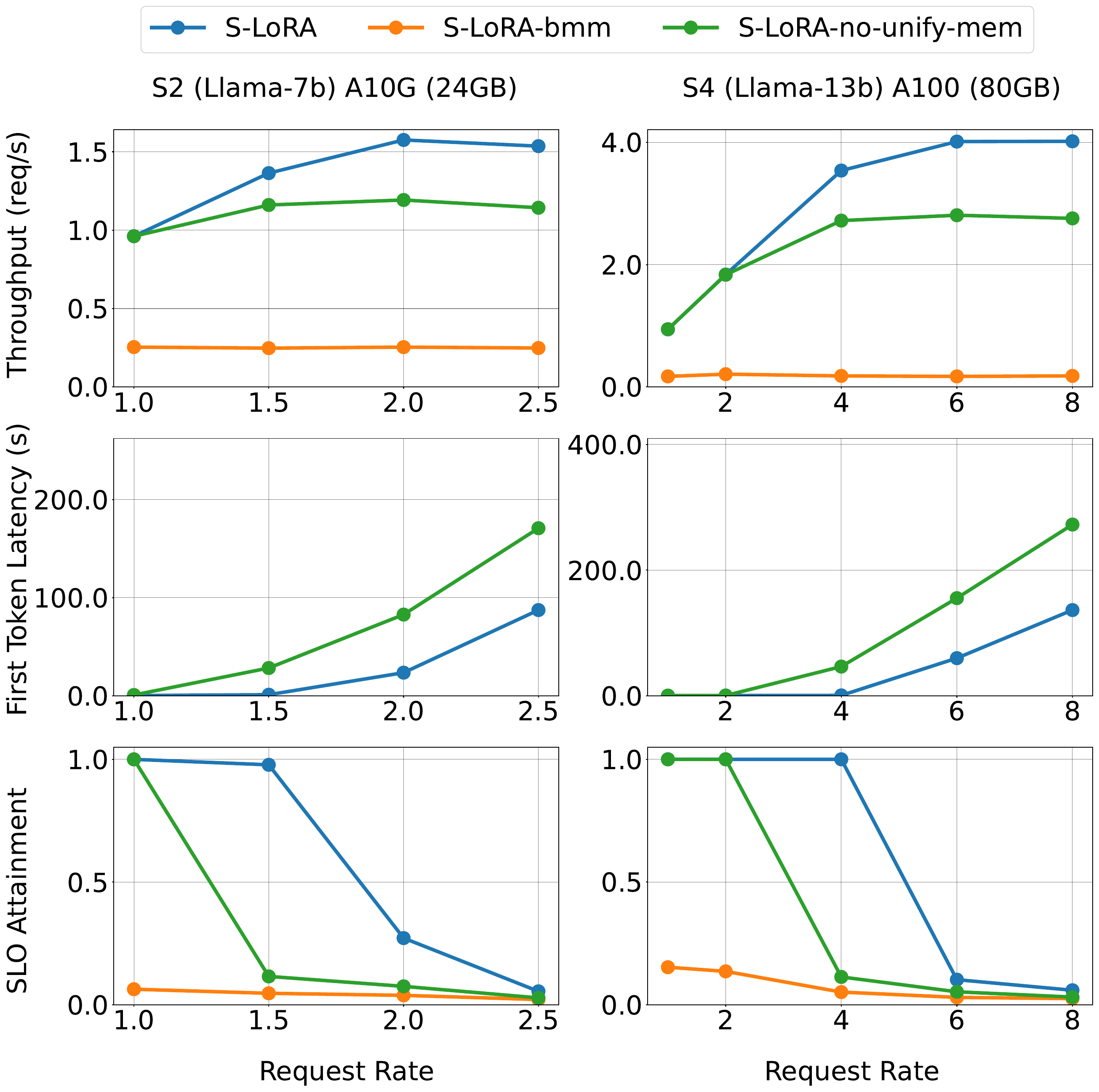}
    \caption{The throughput, first token latency, and SLO attainment of \sys and its variants under different request rates. Note that in both settings the first token latency of \sys-bmm is out of the figure's scope.}
    \label{fig:main_req_rate}
\end{figure}

\autoref{fig:main_req_rate} demonstrates the variation in throughput, first token latency, and SLO attainment relative to the total request rate, revealing a pattern consistent with the aforementioned observations and underscoring the efficacy of our design.

\subsection{End-to-End Results on Real Workloads}

\begin{figure}[ht]
    \centering
    \includegraphics[width=0.48\textwidth]{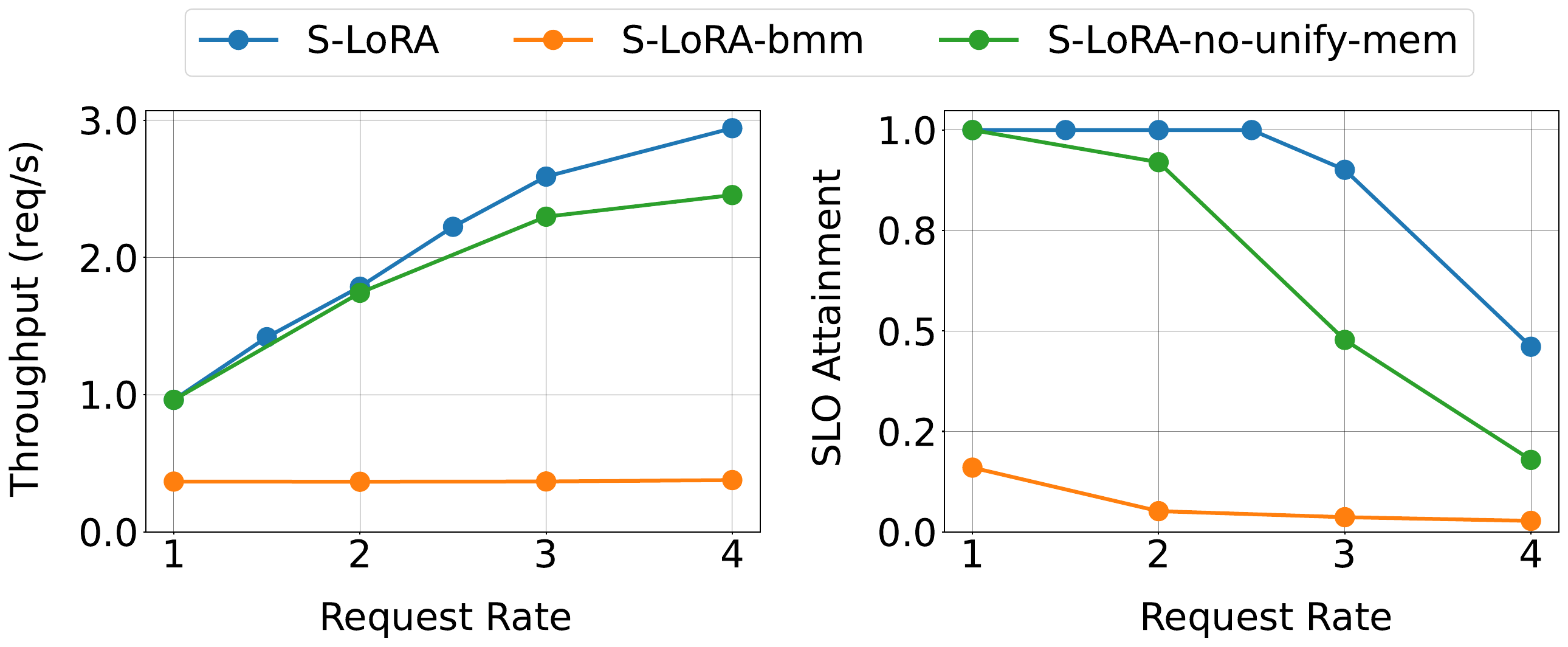}
    \caption{The throughput of \sys and its variants on real workload traces with different request rates. The model and hardware configuration is S2 on an A10G (24GB).}
    \label{fig:real_trace}
    \vspace{-1em}
\end{figure}

\textbf{Real workload trace.}
We construct real-world serving traces by downsampling from the traces of LMSYS Chatbot Arena~\cite{zheng2023judging, zheng2023lmsys}, a website that serves multiple LLMs. The raw log from Arena does not concern LoRA adapters; it focuses on different base models. Nonetheless, we treat the distribution of different base models as if they were the distribution of different adapters of a single base model.
The raw log can be sampled into traces that exhibit varying request rates, denoted as $R$, and durations, represented by $D$. To achieve this, we sample $R\cdot D$ requests from the raw log and rescale the time stamps to fit within the range of $[0, D]$. The number of models $n$ corresponds to the number of adapters.
Furthermore, we set the adapter ranks based on  \autoref{tab:model_setting} with a round-robin method.

Since we are using a real workload trace, there are no parameters such as $\alpha$, $\lambda_i$, or $cv$.
For consistency, we set the duration to 5 minutes.
We adjust the request rate $R$ to study its impact on performance metrics.
In the sampled trace, the average input length is 85 tokens, the average output length is 165 tokens, and the number of adapters is around 26.

\textbf{Results.} \autoref{fig:real_trace} shows the throughput and attainment results, which show a similar pattern to the synthetic workloads. This means the strong performance \sys holds for real world workloads.

\subsection{Multi-GPU Tensor Parallelism}
We test the scalability of our tensor parallelism strategy by running 1) Llama-30B on two A100 (40GB) and four A100 (40GB) GPUs with 10 to 100 adapters; and 2) Llama-70B on two A100 (80GB) and four A100 (80GB) GPUs with 10 adapters. We then report the serving throughputs. 

\begin{figure}[ht]
    \centering
    \includegraphics[width=\columnwidth]{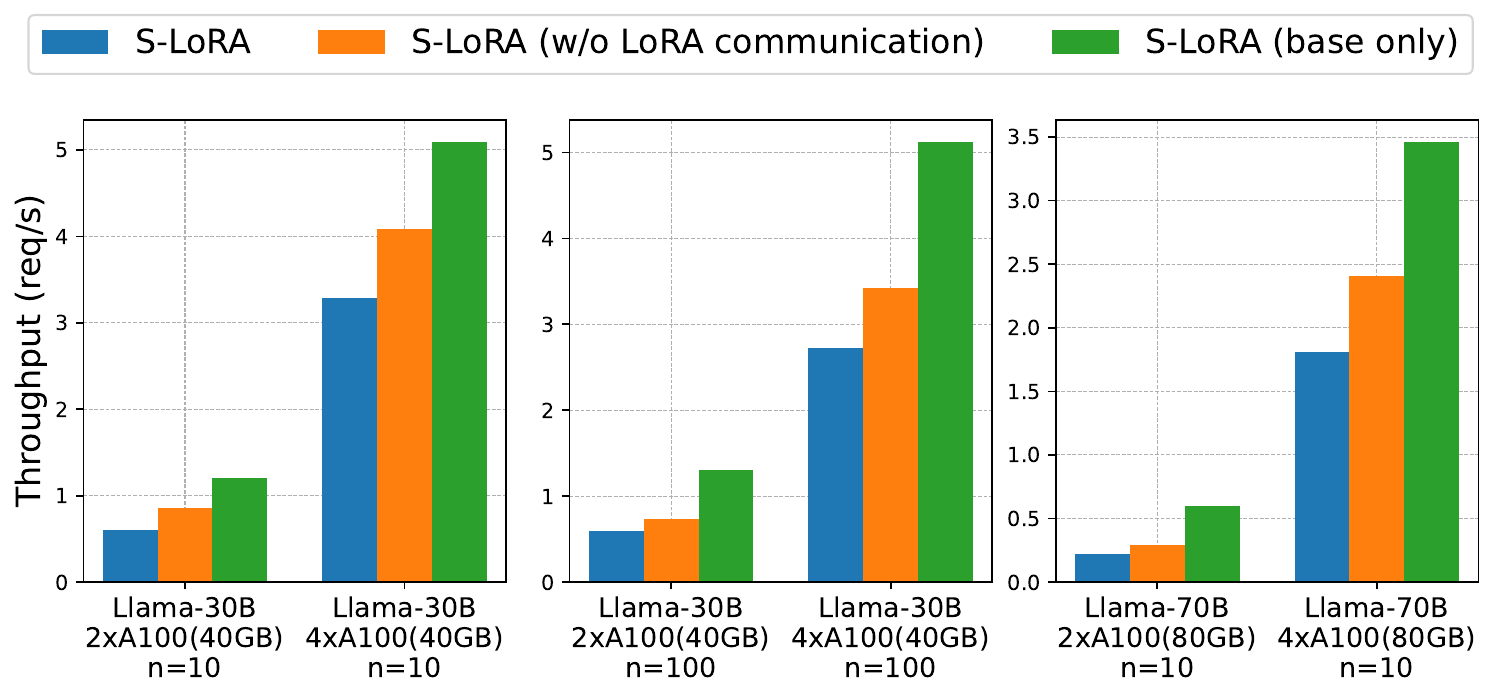}
    \caption{Throughput with tensor parallelism.}
    \label{fig:tp_results}
\end{figure}

As depicted in \autoref{fig:tp_results}, the disparity between \sys with and without LoRA communication is small. This suggests that the added LoRA communication in our strategy has a very small overhead. The figure further reveals that the communication overhead due to LoRA is less than the computational overhead it introduces.
Furthermore, when transitioning from 2 GPUs to 4 GPUs, the serving throughput increases by more than 2 times. This significant increase can be attributed to the fact that the system is predominantly memory-bound in this context. Adding more GPUs alleviates memory constraints, leading to superlinear scaling.
In conclusion, the results verify both the minimal overhead and the scalability of our tensor parallelism strategy.

\subsection{Ablation Study}
\label{subsec:ablation}

\textbf{Merging adapter weights versus computing on-the-fly.}
While \sys does not merge adapter weights and computes LoRA matrices on-the-fly each time, we compare it with an alternative design that merges an adapter with the base model, denoted as \( x(W + AB) \), as proposed in the LoRA paper. This approach involves: 1) Updating the base model with the current adapter weights before each new batch; and 2) Switching to a new adapter if there are too many waiting requests.\footnote{This is different from PEFT. For example, it has continuous batching and PagedAttention, which are not enabled in PEFT.}
This method is efficient for a small number of adapters due to the reduced LoRA computation overhead.

\begin{figure}[ht]
    \centering
    \begin{subfigure}{.32\textwidth}
        \centering
        \includegraphics[width=\textwidth]{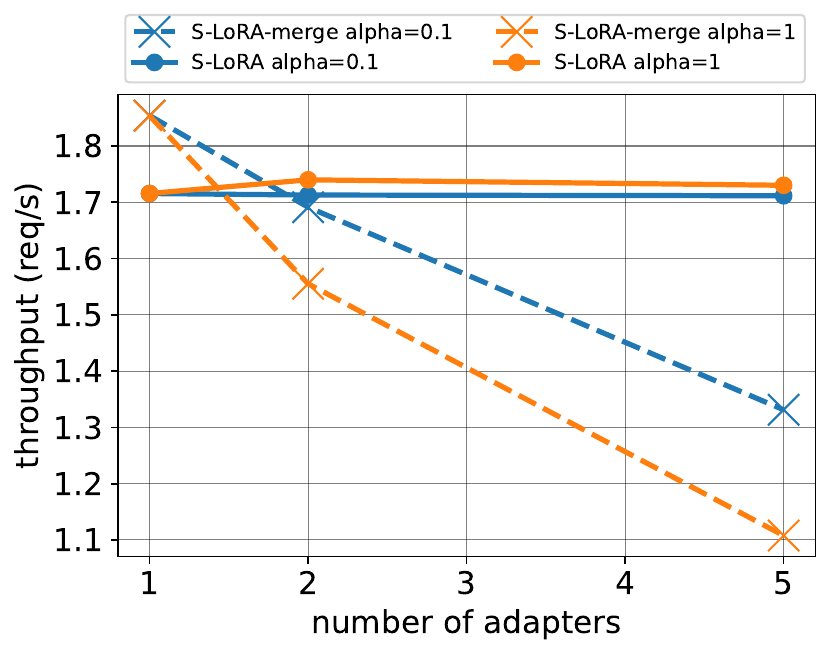}
    \end{subfigure}
    \vspace{-0.5em}
    \caption{Ablation study comparing adapter merging and on-the-fly compute for S2 on A10G (24GB) with different $\alpha$ and number of adapters. The settings for the synthetic workloads are $R = 2, \, cv = 1, \, [I_t, I_u] = [8, 512], \, [O_l, O_u] = [8, 512] $.}
     \label{fig:merge}
\end{figure}

Results in \autoref{fig:merge} demonstrate that with just one adapter, the merging approach outperforms the on-the-fly computation owing to a one-time merging cost. However, its performance declines with more than 2 adapters, primarily because of the time-consuming switch between adapters.
Such switching results in periods of GPU under-utilization. Furthermore, a smaller value of $\alpha$ causes requests to be distributed unevenly across adapters, which in turn reduces batch sizes and overall performance.

\begin{figure}
    \centering
    \includegraphics[width=0.48\textwidth]{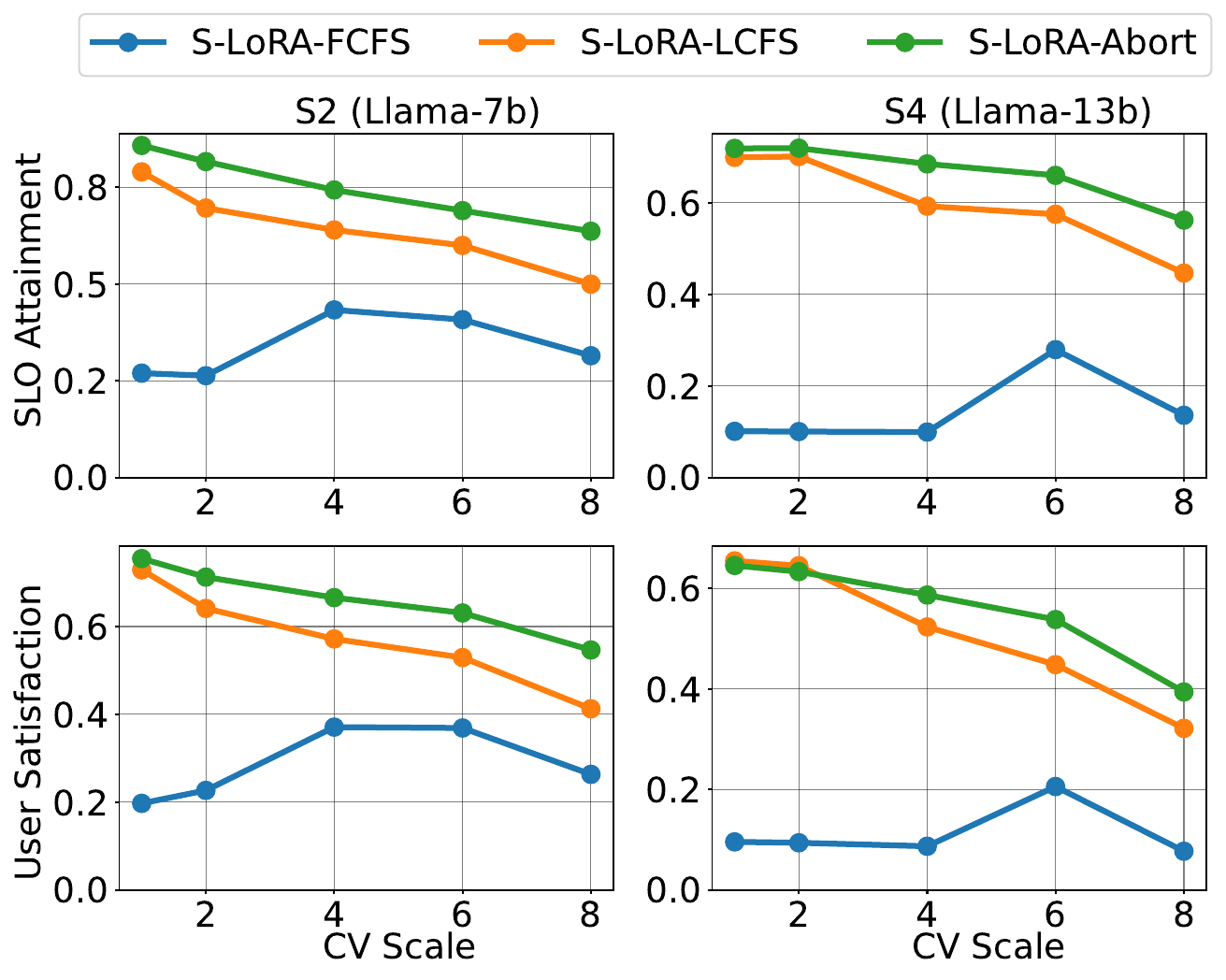}
    \caption{Ablation study for early abort scheduling strategy on A10G-24G (S1) and A100-80G (S4). Other settings follow the description in \autoref{tab:default_trace}.}
    \label{fig:abort}
    \vspace{-0.5em}
\end{figure}

\textbf{Early abort strategy experiments.} We compared \sys's early abort strategy to First Come First Serve (FCFS) and Last Come First Serve (LCFS) for optimizing user satisfaction and SLO attainment. As shown in \autoref{fig:abort}, \sys-Abort outperforms both, especially as $cv$ scales.
FCFS is least effective, often processing requests that have already missed the SLO.
LCFS, similar to a greedy algorithm that only prioritizes the newest requests, works well for small $cv$, but its performance drops with larger $cv$.
\sys-Abort excels as it avoids prioritizing only the newest requests, as detailed in \autoref{sec:admission-detail}.

\section{Related work}
\label{sec:related-work}

\textbf{Optimize LLM serving with system techniques.}
The significance of the transformer architecture has led to the development of many specialized serving systems for it.
These systems use advanced batching mechanisms~\cite{fang2021turbotransformers,yu2022orca},
memory optimizations~\cite{DBLP:conf/icml/0007ZYLRCLRSZ23,kwon2023efficient},
GPU kernel optimizations~\cite{wang2021lightseq,aminabadi2022deepspeed,nvidiaft,dao2023flashattention},
model parallelism~\cite{pope2022efficiently,aminabadi2022deepspeed}, parameter sharing~\cite{zhou2022pets}, and speculative execution~\cite{stern2018blockwise,miao2023specinfer} for efficient serving.
Among them, PetS~\cite{zhou2022pets} is most relevant to ours.
However, PetS only considers the serving for small encoder-only BERT models.
It does not consider generative inference, a very large number of adapters or large models go beyond a single GPU, so it does not address the problems in our settings.

In concurrent work, Punica~\cite{chen2023punica} explored the concept of decomposed computation for the base model and adapters. Some of our CUDA kernels were developed based on the implementation presented in a previous blog post of Punica, with additional support for batching different ranks and non-contiguous memory. Analyzing kernel performance is not the focus of this paper, but it is discussed in Punica. Our work differs from Punica in our novel memory management and tensor parallelism techniques, which have not been covered in any previous work.

\textbf{Optimize LLM serving with algorithm techniques.}
In addition to system-level improvements, inference efficiency can be enhanced using algorithm techniques like quantization~\cite{yao2022zeroquant,dettmers2022gptint,frantar2022gptq,DBLP:conf/icml/XiaoLSWDH23,lin2023awq}, sparsification~\cite{frantar2023massive,zhang2023h} and model architecture improvements~\cite{shazeer2019fast}.
These approaches can reduce memory consumption and accelerate the computation, with a minor compromise in model quality. 
They are complementary to the techniques in this paper.

\textbf{Parameter-efficient fine-tuning.}
Recent work has developed methods for parameter-efficient fine-tuning of large pre-trained language models. These methods show fine-tuning is possible with only a small fraction of tuned parameters.
The state-of-the-art methods include LoRA~\cite{hu2021lora}, Prefix-tuning~\cite{li2021prefix}, P-Tuning~\cite{liu2021p}, Prompt tuning~\cite{liu2023gptunderstand,lester2021power}, AdaLoRA~\cite{zhang2022adaptive}, and $(IA)^3$~\cite{liu2022few}. While our paper focuses on LoRA due to its wide adoption, most techniques can be easily applied to other parameter-efficient fine-tuning methods as well.

\textbf{General purpose model serving systems.}
Over the years, the domain of general model serving has seen significant advancements, Notable systems from earlier research include Clipper~\cite{crankshaw2017clipper}, TensorFlow Serving~\cite{olston2017tensorflow}, Nexus~\cite{shen2019nexus}, InferLine~\cite{crankshaw2020inferline}, and Clockwork~\cite{gujarati2020serving}. These systems delve into topics such as batching, caching, and model placement, catering to both individual and multiple model deployments.
In more recent developments, DVABatch~\cite{cui2022dvabatch}, REEF~\cite{han2022microsecond}, Shepherd~\cite{zhang2023shepherd} and AlpaServe~\cite{li2023alpaserve} have explored the ideas of multi-entry multi-exit batching, preemption, and statistical multiplexing with model parallelism.
Although these systems have made significant contributions, they overlook the auto-regressive characteristics and parameter-efficient adapters in LLM serving, leading to potential optimization gaps.

\section{Conclusion}
\label{sec:conslusion}

We present \sys, a system capable of serving thousands of LoRA adapters from a single machine with much higher throughput compared to existing systems.
\sys is made possible by our innovative design of the unified memory pool, tensor parallelism strategy, adapter batching, and CUDA kernels. 
\sys enables large-scale, customized fine-tuning services essential for deploying models tailored to diverse requirements.
Future extensions of \sys will encompass support for additional adapter methods, enhanced fused kernels, and the use of multiple CUDA streams to parallelize base model and LoRA computations.

\section*{Acknowledgment}
This research was supported by gifts from Anyscale, Astronomer, Google, IBM, Intel, Lacework, Microsoft, Mohamed Bin Zayed University of Artificial Intelligence, Samsung SDS, Uber, and VMware.
Ying is partly supported by the Stanford Center for Automated Reasoning.
We thank Clark Barrett for academic advising and funding support.
We also thank Yonghao Zhuang and Lisa Dunlap for their helpful discussions and feedback.

\bibliography{reference}
\bibliographystyle{mlsys2024}
\clearpage

\appendix
\newpage
\section{Additional Experiment Results}
\label{sec:full_exp}

\subsection{Analysis of PEFT}
In our evaluation of PEFT, several key observations were discerned. First, the lack of KV cache support makes the maximal batch size of PEFT much smaller compared to \sys. For instance, in A10G S1, \sys can accommodate a maximal batch size of 30, while PEFT can only accommodate a maximal batch size of 6. Secondly, the lack of continuous batching support makes shorter requests wait for longer requests in a batch. These two factors together result in the low throughput of PEFT even when there is only one adapter. When there are more adapters, the lack of batching support across different adapters makes the throughput even lower, resulting in only 0.17 request/second performance in the largest number of adapters we test. As another result, the average latency explodes because the request rate is far beyond the maximal capacity of the PEFT system. In Table~\ref{tab:peft_req_rate}, we show that even in the lowest request rate we test, PEFT fails to process with a low latency.

\begin{table}[ht]
\centering
\resizebox{1\columnwidth}{!}{
\begin{tabular}{c|ccc}
\toprule
num adapters & throughput & avg. latency & avg. attainment \\
\midrule
1 & 0.26 & 1021.86 & 0.0\\
20 & 0.23 & 1178.52 & 0.0\\
50 & 0.22 & 1293.97 & 0.0\\
100 & 0.20 & 1421.16 & 0.0\\
200 & 0.17 & 1609.50 & 0.0\\

\bottomrule
\end{tabular}}
\caption{PEFT results on the synthetic workload S1 against number of adapters.}
\label{tab:peft_num_adapters}
\end{table}

\begin{table}[ht]
\centering
\resizebox{1\columnwidth}{!}{
\begin{tabular}{c|ccc}
\toprule
req rate & throughput & avg. latency & avg. attainment \\
\midrule
1 & 0.11 & 1165.46 & 0.0\\
1.5 & 0.13 & 1398.56 & 0.0\\
2 & 0.17 & 1614.37 & 0.0\\
2.5 & 0.18 & 1904.73 & 0.0\\

\bottomrule
\end{tabular}}
\caption{PEFT results on the synthetic workload S1 against request rate.}
\label{tab:peft_req_rate}
\end{table}

\subsection{Experiments for adapter clustering.}
\label{sec:ablation_clustering}
\begin{figure}[ht]
    \centering
    \includegraphics[width=0.48\textwidth]{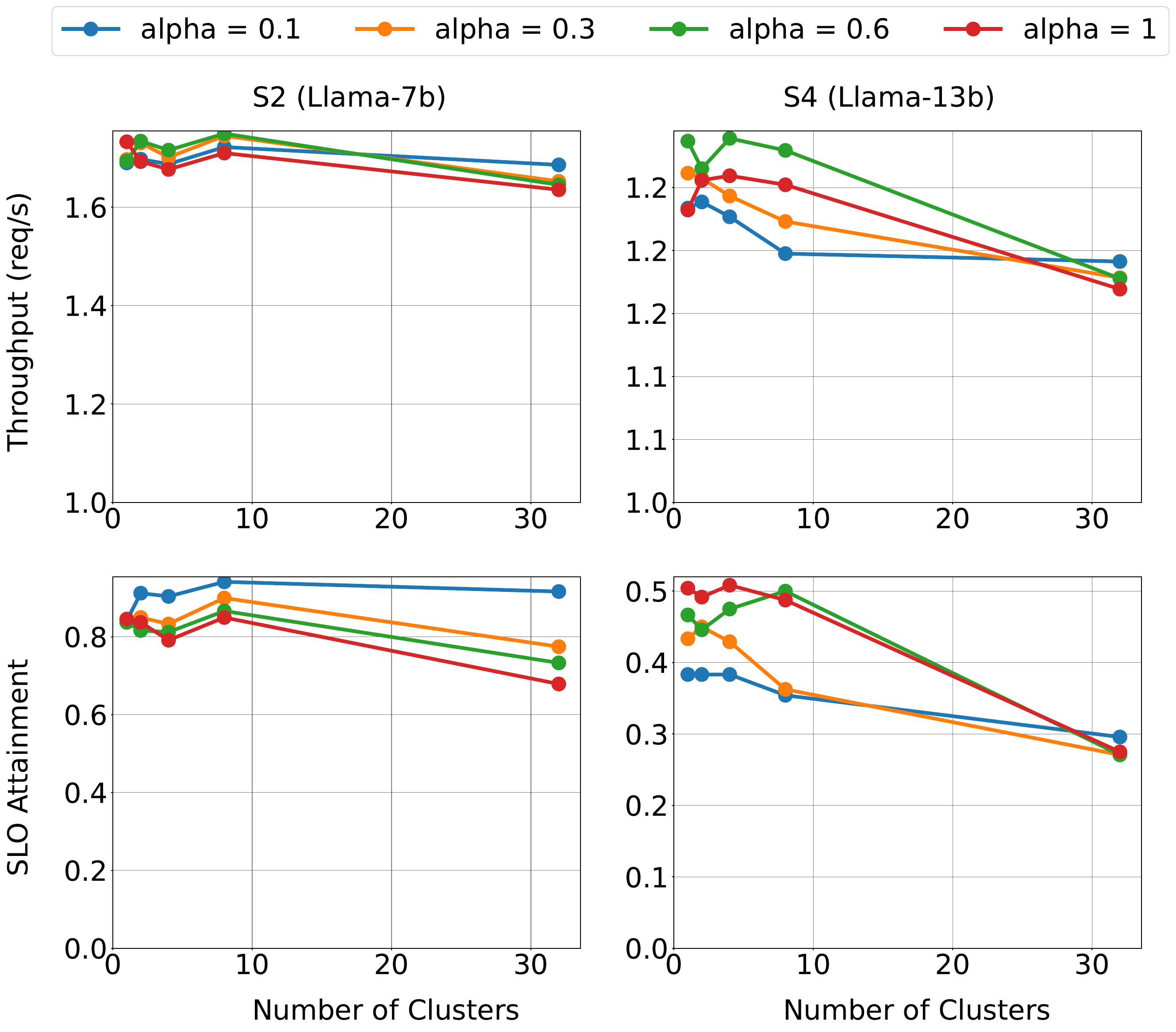}
    \caption{Ablation study for different number of clusters on A100 (40GB) with different $\alpha$. The settings for the synthetic workload trace are $n = 32, \alpha = [0.1, 0.3, 0.6, 1], \, R = 2, \, cv = 1, \, [I_t, I_u] = [8, 512], \, [O_l, O_u] = [8, 512] $}
    \label{fig:ablation_cluster_size_alpha}
\end{figure}

\begin{figure}[ht]
    \centering
    \includegraphics[width=0.48\textwidth]{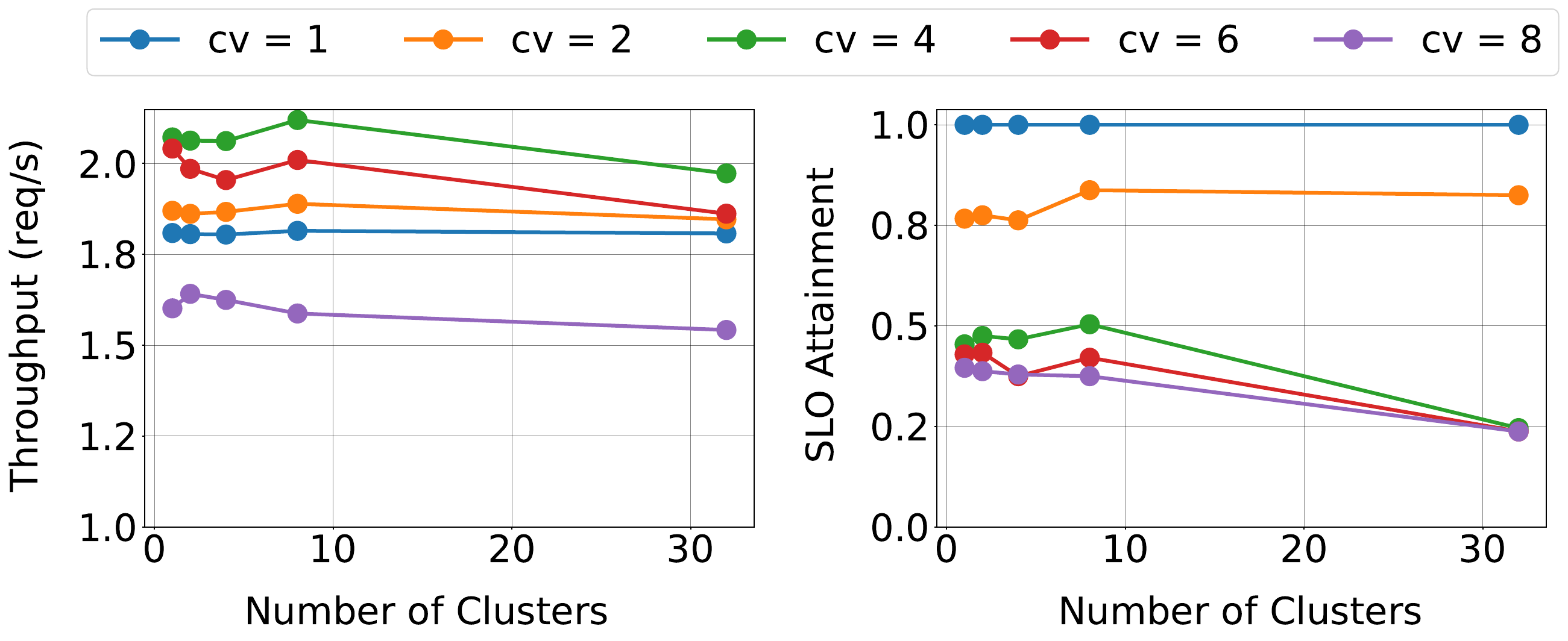}
    \caption{Ablation study for different number of clusters on S2 (Llama-7b) A100 (80GB) with different $cv$. The settings for the synthetic workload trace are $n = 32, \alpha = 1, \, R = 2, \, cv = [1, 2, 4, 6, 8], \, [I_t, I_u] = [8, 512], \, [O_l, O_u] = [8, 512] $}
    \label{fig:ablation_cluster_size_cv}
\end{figure}

We implement a straightforward adapter clustering algorithm.
Let parameter $d$ be the number of adapters in a batch.
In addition to the FCFS order (or early abort order if turned on), if the number of adapters reaches $d$, we will prioritize the requests that have their adapter already in the batch.
But if the requests from the $d$ adapters cannot fill all the space for a running batch, we allow other requests to be added.
We run some additional experiments to study how the number of clusters impacts throughput and SLO attainment.
We call $d$ as the number of clusters in the figure.
As shown in \autoref{fig:ablation_cluster_size_alpha} and \autoref{fig:ablation_cluster_size_cv},
the impact is not significant but observable, especially for larger $\alpha$ and $cv$.
Generally, a small $d$ can result in better performance.
The small fluctuation for small $d$'s may be because of the scheduler overhead and random noise.

\section{Admission Control in \sys}
\label{sec:admission-detail}
Traditional admission control usually assumes a hard threshold for the delay in the service~\citep{jamin1993admission, vin1994statistical, naghshineh1996distributed}, and controls the total number of violations of delay. Here for LoRA serving, we assume a soft threshold characterized by the user's reward function.
For illustration purposes, let the arrival time of the requests be integers, and assume that we process one query in each time period of length $1$. Let $Q = \{q_1, q_2, \cdots q_n\}$ be the request queue in the ascending order of the arrival time, and $l$ be the desired number of served requests. We quantify the user's satisfaction with a reward function  $r:\mathbb{R}^+\mapsto[0, 1]$ that maps the first token latency of a request to a scalar in between $[0, 1]$, where $0$ represents the user losing patience and giving up the query, and $1$ represents the user is completely satisfied with the latency. Let $t_i$ be the latency of serving the request $q_i$ in the queue $Q$. Then we aim to solve the following constrained optimization:
\begin{align}
    \max & \sum_{i=1}^n r(t_i) \label{eq:constrained} \\
     \text{ s.t. } & \mathbbm{1}(r(t_i) > 0 ) = l\nonumber.
\end{align}
We show that when the derivative of reward is non-increasing, the optimal solution to the above constrained optimization problem is to serve the most recent $l$ elements $q_{n-l+1}, q_{n-l+2}, \cdots, q_n$ in order.
\begin{theorem}\label{thm:admission_control}
    Assume that $r'(t) \leq 0$ for any $t\in\mathbb{R}^+$. The optimal solution to Equation (\ref{eq:constrained}) is to serve the most recent $l$ elements $q_{n-l+1}, q_{n-l+2}, \cdots, q_n$ in order.
\end{theorem}
The proof is deferred to Appendix~\ref{proof:admission_control}.  In practice, for a given request queue, we can estimate the largest possible number of requests to be served in SLO as $l$. Then we take the most recent $l$ elements for serving.
Such an $l$ can be approximated by simulating a First-Come-First-Serve (FCFS) strategy, which is optimized to serve requests as many as possible.

In \sys, the scenario is more complicated because of the 
heterogeneity and unpredictability of the sequence length.
As an approximation, we implement a heuristic as follows.
The high-level scheduling is that we will fetch a minibatch of new requests to be added into the running batch every several decode step.
From the history, we use the moving average to estimate a current request rate $R_1$ measured in how many requests will be added to the waiting queue per period of fetching new requests. We also use the moving average to estimate the number of new requests $R_2$ that can be added to the running batch for a period.
Let $rt_i$ be the coming time of request $r_i$, $ct$ be the current time, $tl_{max}$ be the maximum allowed first token latency to meet the SLO and $l_{prefill}$ be the maximum prefill latency for a minibatch in history. Each time we generate the new minibatch, we will first abort the requests $R = \{r_k \mid ct - rt_k + l_{prefill} > tl_{max}\}$. Requests in $R$ are highly likely to miss the SLO even if they get scheduled immediately due to the high prefill latency. Then if $R_1 > R_2$, which means the system is temporarily overloaded, we will fetch the newest requests into the minibatch. If $R_1 \leq R_2$, the waiting queue will be shortened if the trend continues. In this case, we will choose from the earliest.

\subsection{Proof of Theorem~\ref{thm:admission_control}}\label{proof:admission_control}

We first prove that for any admission control strategy that serves $l$ elements, one can always find another admission control strategy that serves the most recent $l$ elements with a larger cumulative reward. 

Assume that  we serve $l$ elements $q_{s_1}, q_{s_2}, \cdots, q_{s_l}$ in the $l$ timesteps. Assume without loss of generality that $q_{s_1}$ is not among the most recent $l$ elements, and assume that the $k$-th element is not served with $k\in[n-l, n]$. By definition we know that $s_1<k$. Now at the time of serving $q_{s_1}$, we serve $q_k$ rather than $q_{s_1}$, and keep the rest of the choices in other time steps same. In this case, the number of served queries remains the same. On the other hand, we know that the latency satisfies $t_{s_1}> t_{k}$ since the $k$-th element is more recent. This gives that 
\begin{align*}
    r(t_{s_1}) < r(t_k).
\end{align*}
Since the reward for other elements does not change,  the total reward is increased while the constraint is still satisfied. By repeating the operations until all the elements served are the most recent $l$ elements, we prove that claim. 

Next, we prove that serving the most recent $l$ elements in order of $q_{n-l+1}, q_{n-l+2}, \cdots, q_{n}$ is optimal. For any $i, j\in[n-l+1, n]$, we assume that $i<j$ and $j$ is first served at time $t_1$ while $i$ is served at time $t_2$ with $t_1<t_2$. Let $t_i^a$, $t_j^a$ be the arrival time of $i, j$. The reward for serving $i, j$ in this case becomes
\begin{align*}
    r(t_2 - t_i^a) + r(t_1 - t_j^a).
\end{align*}
Now we show that by swapping the time of serving $i, j$, the reward does not decrease. This is equivalent to showing that
\begin{align*}
   r(t_1 - t_i^a) + r(t_2 - t_j^a) \geq  r(t_2 - t_i^a) + r(t_1 - t_j^a).
\end{align*}
Rearranging the above equation, we know that it is equivalent to
\begin{align*}
   \frac{r(t_1 - t_i^a) - r(t_2 - t_i^a)}{t_1 - t_2}    \leq    \frac{ r(t_1 - t_j^a) - r(t_2 - t_j^a)}{t_1-t_2}.
\end{align*}
This is true due to the concavity of the reward function, thus finishing the proof.

\end{document}